\documentclass{article}

\usepackage[preprint]{corl_2026} 

\usepackage[table]{xcolor}

\definecolor{lightblue}{HTML}{E6F2FF}
\newcommand{\greenNum}[1]{\cellcolor{lightblue}{#1}}

\definecolor{lightpurple}{HTML}{F2E6FF}
\newcommand{\purpleNum}[1]{\cellcolor{lightpurple}{#1}}

\usepackage{times}

\usepackage{multicol}
\usepackage[font=small]{caption}
\usepackage[dvipsnames,table]{xcolor}
\usepackage{xurl}
\usepackage{array}
\usepackage{colortbl}
\usepackage{tabularx}
\usepackage{amsmath}
\usepackage{comment}
\usepackage{amssymb}
\usepackage{algorithm}

\usepackage{algpseudocode} %
\usepackage{enumitem}
\usepackage{xcolor}
\usepackage{xspace}
\usepackage{graphicx}
\usepackage{subcaption}
\usepackage{booktabs} 
\usepackage{makecell}
\usepackage{multirow}
\usepackage{wrapfig}
\usepackage{array}
\newcolumntype{C}[1]{>{\centering\arraybackslash}p{#1}}
\newcommand{\our}{B-spline policy\xspace}
\newcommand{\pickcube}{Cube Picking\xspace}
\newcommand{\cleantable}{Table Cleaning\xspace}
\newcommand{\stackunstack}{Speed Stacking\xspace}

\newcommand{\cellSRtime}[2]{\makecell{#1\\#2}}

\newcommand{\cellMetric}{\makecell{\textit{Success rate}\\\textit{Avg. time}}}

\newcommand{\pushtcellMetric}{\makecell{\textit{Avg. score}\\\textit{Avg. time}}}

\title{B-spline Policy: Accelerating Manipulation Policies via B-spline Action Representations}

\author{
{\bfseries
Xiaoshen Han$^{1}$\thanks{Equal contribution.}
\quad
Haoyu Xiong$^{2}$\footnotemark[1]}\\[0.5em]
{\bfseries
Haonan Chen$^{1}$ \quad
Chaoqi Liu$^{1}$ \quad
Antonio Torralba$^{2}$ \quad
Yuke Zhu$^{3}$ \quad
Yilun Du$^{1}$}\\[0.5em]
{\normalfont
$^{1}$Harvard \quad
$^{2}$MIT \quad
$^{3}$UT Austin}
}
\begin{document}
\maketitle

\vspace{-20pt}
\begin{abstract} In this work, we present B-spline Policy (BSP), an action representation designed for accelerating robot manipulation policies. Rather than predicting discrete-time action chunks, BSP parameterizes actions as continuous B-spline curves defined by a set of knots and control points. This representation yields smooth, time-continuous trajectories that can be temporally scaled and executed by low-level controllers at higher frequencies and speeds. We show that B-spline–parameterized actions can be seamlessly integrated into standard policy learning pipelines by directly predicting B-spline parameters. Experiments on simulated and real-world tasks demonstrate that BSP significantly reduces task completion time, achieving substantial improvements over baseline methods while maintaining strong success rates.
More results: \href{https://b-spline-policy.github.io/}{B-spline-policy.github.io}
\end{abstract}

\keywords{Fast manipulation; Visuomotor policy speedup; Action representation} 

\section{Introduction}
\begin{wrapfigure}{r}{0.48\textwidth}
  \centering
  \vspace{-18pt}
  \includegraphics[width=0.46\textwidth]{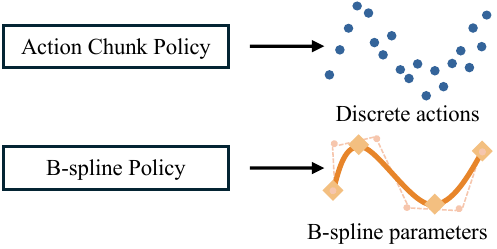}
  \caption{
  \textbf{\our.} Rather than predicting a chunk of discrete-time actions, \our parameterizes actions as continuous B-spline\protect\footnotemark curves. This continuous representation enables temporal rescaling at execution time, allowing the policy to execute faster and more smoothly.
  }
  \label{fig:teaser-fig}
  \vspace{-10pt}
\end{wrapfigure}
\footnotetext{A B-spline defines a smooth curve by a set of knots and control points. Instead of drawing straight lines between waypoints, a B-spline creates a smooth path that follows their overall trend. Think of it like bending a flexible ruler near waypoints instead of connecting them with sharp corners.}
Robotic manipulation via visuomotor policy learning has made remarkable progress in recent years~\cite{pi_05,gr1,act,dp}. 
Yet, despite these advances, task execution speed remains a major bottleneck. In everyday manipulation tasks such as folding a T-shirt, humans typically complete the task in roughly 10 seconds, whereas even state-of-the-art robotic systems often require close to a minute~\cite{tshirt, pi_05, black2024pi_0}. This gap exposes a fundamental and largely unresolved challenge in visuomotor policy learning—\emph{efficiency}: how can robots execute complex manipulation tasks quickly, rather than merely complete them successfully?


A key source of inefficiency in modern visuomotor policies lies in the design of action chunking~\cite{dp,act}. Most existing approaches parameterize actions as fixed-length chunks sampled uniformly over time. While such chunking can stabilize long-horizon prediction and improve learning performance~\cite{mobile-aloha}, it also introduces inherent limitations: 
\vspace{-5pt}

\begin{itemize}[leftmargin=3mm]
    \item \textit{Uniform temporal resolution.} 
    Fixed-length chunks assume that every phase of a task requires identical control granularity. In practice, manipulation tasks exhibit strongly non-uniform temporal structure: fast motions such as reaching can be executed rapidly with coarse actions, whereas contact-rich phases such as grasping, alignment, and insertion demand fine-grained precision. Uniform chunking fundamentally misallocates representational capacity across time.
\end{itemize}
\vspace{-5pt}

\begin{itemize}[leftmargin=3mm]
    \item \textit{Chunk discontinuities.} 
    Stitching together independently predicted action chunks at inference time can introduce discontinuities at chunk boundaries. While such boundary mismatches may be tolerable in low-speed or quasi-static settings, they become highly problematic during high-speed execution, where even small errors can lead to tracking failures or complete task failure.
\end{itemize}


\vspace{-5pt}
These limitations motivate a paradigm shift: representing robot actions as \emph{\textbf{continuous curves}} rather than discrete waypoint chunks. Such a representation is used in classical motion planning~\cite{to-pnp-with-bsplines,to-via-bsplines,bsplines-based-trajectory-design,bspline-on-gcs} and computer graphics~\cite{bspline-in-cad-cg,intro-bspline-graphics}, where spline-based representations are widely favored for being smooth, compact, and mathematically tractable. Among spline families, B-splines are especially attractive for visuomotor control: they are smooth by construction, provide compact parameterizations~\cite{splines-guide,calculating-with-B-splines,fitting-bspline}, and can approximate continuous functions arbitrarily well given sufficient knots~\cite{approximation-bspilnes,splines-guide}.

In this work, we propose \our, a visuomotor policy that directly outputs {\it continuous action curves}.
As illustrated in Fig.~\ref{fig:teaser-fig}, rather than predicting discrete-time action chunks, \our outputs B-spline parameters that define an action curve in continuous-time. This shift from discrete action chunks to continuous action curves provides several key advantages:

\vspace{-9pt}

\begin{itemize}[leftmargin=3mm]
    \item \textit{Acceleration via temporal rescaling.} 
    Each predicted B-spline segment can be sampled at arbitrary control frequencies and retimed dynamically by scaling the temporal mapping. This allows the robot to execute the identical trajectory at significantly higher speeds without sacrificing fidelity.
\end{itemize}

\vspace{-10pt}

\begin{itemize}[leftmargin=3mm]
    \item \textit{Inference-time segment alignment.} 
    To ensure stable accelerated execution over the duration of a task, we introduce an inference-time segment alignment mechanism that enforces smooth transitions between the predicted B-spline segments.
\end{itemize}

\vspace{-10pt}

\begin{itemize}[leftmargin=3mm]
    \item \textit{Plug-and-play integration.} 
    \our requires minimal modifications to existing imitation learning methods, serving as a drop-in replacement for standard action chunk prediction.
\end{itemize}

We evaluate \our across both simulated and real-world manipulation environments. Our real-world setup includes three tasks spanning distinct robotic challenges: precise object manipulation, long-horizon tabletop rearrangement, and bimanual, contact-rich interaction. Integrated into both Diffusion Policy and ACT-style backbones, \our consistently reduces task completion time while maintaining or improving success rates. Our results demonstrate that \our yields substantially smoother executions, and that our alignment mechanism is critical for stable, high-speed control.

Our contributions are threefold: \textbf{(1)} we introduce a B‑spline–based action representation with adaptive temporal resolution for visuomotor imitation learning; \textbf{(2)} we propose segment alignment to enable smooth transitions during accelerated policy execution; and \textbf{(3)} We demonstrate across diverse real-world and simulated tasks that \our substantially reduces task completion times compared to standard action chunking baselines while preserving task success.


\vspace{-7pt}
\section{Related work}
\vspace{-3pt}

\textbf{Visuomotor Policy Acceleration.}
High-speed manipulation has been extensively studied in classical robotics, especially in motion planning and trajectory optimization~\cite{cmu-motion-plan-jeff}.
However, these techniques are not directly applicable to visuomotor policy learning, where actions must be predicted end-to-end from high-dimensional sensory observations.
In policy learning, robot manipulation policies often remain slow, yet performance is still primarily evaluated by success rate~\cite{pi_05, bidex, via, dp, manip_as_in_sim}, with comparatively less attention to task completion time.
Recent works~\cite{rtc, demospeedup, arachchige2025sail} have begun to address this gap. For example, DemoSpeedup~\cite{demospeedup} accelerates imitation policies by using action entropy to identify which parts of demonstrations can be safely downsampled, while SAIL~\cite{arachchige2025sail} frames this problem as a full-stack engineering involving policy consistency, robot dynamics, and latency-aware scheduling.
In contrast, our work identifies action chunking as a key algorithmic bottleneck for fast visuomotor control. We therefore introduce B-spline Policy (BSP), which replaces fixed discrete action chunks with continuous B-spline curves, enabling smooth and fast manipulation.

\textbf{B-spline Action Representations.} 
Discrete-time action chunking is widely used in visuomotor policy learning~\cite{act, dp, bidex, via, factor-dp, pi_05, re3sim}.
While fixed-length action chunking has been shown to stabilize long-horizon visuomotor learning~\cite{block2023provbc, mobile-aloha}, it imposes a uniform temporal resolution that fails to capture the inherently non-uniform structure of manipulation tasks.
%
%
%
In classical motion planning, actions are often parameterized as continuous-time curves, such as B'ezier curves~\cite{bezier-for-robot-motion-plan} and splines~\cite{to-via-bsplines, bsplines-based-trajectory-design, bspline-on-gcs}. These representations provide intrinsic smoothness and analytic derivatives~\cite{grotjahn2002model, an1987experimental}. Motivated by these properties, we adopt B-splines as a continuous action representation for accelerating learned visuomotor policies.
%
%
%
Closely related to our work, BEAST~\cite{beast} proposes a B-spline encoded action tokenizer for efficient action sequence modeling and decoding in VLA policies. DMPs~\cite{ijspeert2013dynamical} represent robot motions as stable, temporally scalable dynamical systems with learnable forcing terms.
In contrast, BSP uses B-splines not as a trajectory encoding or motion primitive, but as a control-oriented continuous action representation and execution pipeline for accelerating end-to-end visuomotor policies.


\section{B-spline Action Representations}

In this section, we formulate the B-spline action representation, which shifts visuomotor control from discrete-time action chunk prediction to continuous and smooth B-spline trajectory generation.

\begin{wrapfigure}{r}{0.48\textwidth}
  \vspace{-15pt}
  \centering
  \includegraphics[width=0.5\textwidth]{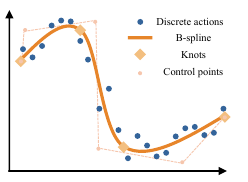}
\caption{\textbf{Adaptive B-spline fitting for demonstration trajectories.}
Given a discrete-time demonstration trajectory, we convert it into a continuous B-spline representation by adaptively selecting knots and estimating control points under a bounded reconstruction error.}
\label{fig:data-fitting}
  \vspace{-20pt}
\end{wrapfigure}

\paragraph{Discrete Action Chunking.} 
We focus on learning a robot policy $\pi(a \mid o)$ from demonstration datasets. The policy maps proprioceptive and visual observations $o_t$ to an \emph{action chunk} 
$\boldsymbol{a}_t = [\boldsymbol a_t,\boldsymbol a_{t+1}, \ldots, \boldsymbol a_{t+T_a}],$
where $T_a$ denotes the prediction horizon.
This action chunking paradigm is widely adopted in modern robot policies~\cite{dp, act, black2024pi_0}, where actions typically correspond to joint positions or end-effector poses.

While predicting action chunks stablizes long-horizon consistency~\cite{dp, block2023provbc, zhang2025actionchunkingexploratorydata}, it also introduces two key limitations for fast manipulation. First, fixed-length chunks impose a uniform temporal resolution across the entire task. Second, independently predicted chunks must be stitched together at inference time, which can create boundary discontinuities that become especially problematic during high-speed execution.

\paragraph{Continuous Action Curves.} To address these limitations, we parameterize the action space as a continuous B-spline trajectory, rather than a discrete sequence of waypoints.

Formally, a B-spline trajectory is defined by a knot vector $U$ and a set of control points $\mathbf{C} = \{c_0, \ldots, c_N\}$. The continuous action at normalized time $u$ is given by
\begin{equation}
    \mathbf{a}(u) = \sum_{i=0}^{N} N_{i,p}(u) \cdot c_i .
\end{equation}
where $N_{i,p}(u)$ is the $i$-th B-spline basis function of degree $p$. In our experiments, we use cubic B-splines, which provide smooth trajectories with continuous velocity and acceleration profiles. This formulation turns the policy output from a fixed list of discrete waypoints into the parameters of a continuous trajectory. The predicted curve can be sampled at any desired control frequency and executed by the low-level controller as a dense stream of commands.

B-splines provide several properties that are particularly useful for fast robot manipulation:
\newline
\textbf{\textit{1) Temporal flexibility}}. Because the action trajectory is defined continuously, the same curve can be sampled at a higher control frequency without changing the policy. This allows the robot to receive dense low-level commands even when the policy runs at a lower inference rate.

\textbf{\textit{2) Temporal rescaling}}. Given a predicted trajectory $\mathbf{a}(u)$, we can execute it faster by scaling the mapping between real time and the curve parameter. For a speedup factor $n$, the robot follows the same action trajectory while traversing the curve more quickly:
$\mathbf{a}_{\mathrm{exec}}(t) = \mathbf{a}(n t)$.
This provides a mechanism for accelerating a learned policy without retraining it for each target execution speed.

\textbf{\textit{3) Intrinsic smoothness}}. B-splines are smooth by construction. Unlike discrete action chunks, which may contain sharp changes between adjacent predictions, a B-spline trajectory produces smooth interpolated actions. This smoothness is especially important under high-speed execution, where abrupt changes in commanded poses can lead to controller tracking errors, overshoot, or task failure.

\textbf{\textit{4) Local error isolation}}. each control point only affects a limited portion of the trajectory. This locality makes the representation stable and compact. Errors in one control point do not globally distort the entire predicted action sequence.


\begin{figure*}[t]
    \centering
    \includegraphics[width=0.98\linewidth]{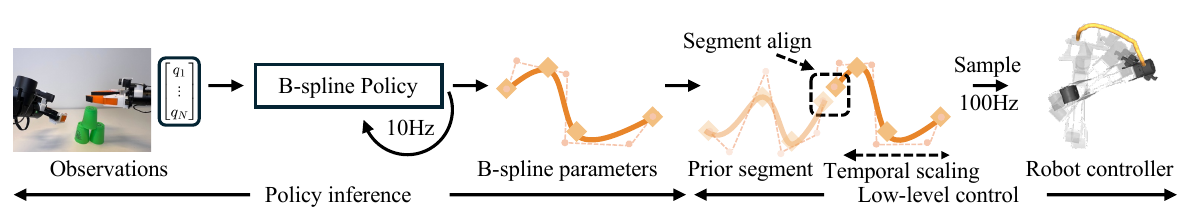}
    \caption{\textbf{\our inference overview.}
    At each inference step, the policy takes images and proprioception as the input and outputs future B-spline segment parameters.
    At the low-level control stage, we align each successive B-spline segment with the prior one. The predicted B-spline segment can be temporally scaled to adjust execution speed. Actions are then sampled at a higher frequency and sent to the robot controller.}
    \label{fig:method}
    \vspace{-12pt}
\end{figure*}



B-splines offer a natural continuous representation for robot actions, however, using them in a visuomotor policy requires several design choices. In particular, we need to convert discrete demonstrations into B-spline targets, represent non-uniform knots with a fixed-size policy output, and execute consecutive spline segments smoothly. We describe these components in the next section.

\section{B-spline Policy}
\label{sec:model-train}

In this section, we describe how to build a visuomotor policy with B-spline actions and convert discrete demonstrations into fixed-size B-spline targets.

\noindent\textbf{Adaptive B-spline Fitting from Demonstrations.}
%
%
%
%
%
%
%
To train \our, we first convert discrete demonstration trajectories into continuous B-spline trajectories.
Given a sequence of demonstration actions, we fit a degree-$p$ B-spline that approximates the original trajectory within an error tolerance. We use an adaptive fitting procedure based on the classical FITPACK strategy~\cite{fitting-bspline,dierckx1995curve,dierckx1982algorithms}, which iteratively inserts knots into trajectory regions with the largest reconstruction error.
\newline
As shown in Fig. \ref{fig:data-fitting}, this adaptive knot placement allocates representation capacity where it is needed most: fewer knots are allocated to smooth regions, while density increases in high-curvature segments. As a result, the fitted B-spline provides a compact continuous representation of the demonstration trajectory while maintaining bounded approximation error, as summarized in Algorithm \ref{alg:spline-compress}.

\begin{wrapfigure}{r}{0.55\textwidth}
\vspace{-20pt}
\centering
\begin{minipage}{0.55\textwidth}
\begin{algorithm}[H]
\caption{Adaptive B-spline Curve Fitting}
\label{alg:spline-compress}
\small
\textbf{Input:} samples $\{(t_i, \mathbf{a}_i)\}_{i=1}^N$, tolerance $\varepsilon$, max knots $|U|_{\max}$ 
\begin{enumerate}[leftmargin=1.2em,itemsep=0pt,topsep=2pt]
\item Initialize $U \gets \{t_1, t_N\}$.
\item \textbf{while} $|U| < |U|_{\max}$:
  \begin{enumerate}[leftmargin=1.2em,itemsep=0pt,topsep=0pt]
  \item Fit degree-$p$ B-spline $\mathbf{a}(\cdot)$ with knots $U$.
  \item $E \gets \max_i \|\mathbf{a}(t_i) - \mathbf{a}_i\|$.
  \item \textbf{if} $E \le \varepsilon$, \textbf{break}.
  \item Insert one knot via FITPACK criterion.
  \end{enumerate}
\item \textbf{return} $U$, $\mathbf{a}(\cdot)$.
\end{enumerate}
\textbf{Output:} knot vector $U$, fitted trajectory $\mathbf{a}(\cdot)$

\end{algorithm}
\end{minipage}
\vspace{-1em}
\end{wrapfigure}

\noindent\textbf{Fixing the Policy Output Size.} After fitting B-splines to the demonstrations, we train a policy to predict B-spline parameters from observations. Directly predicting a full-trajectory spline is inefficient, so the policy outputs the \emph{next} B-spline segment at each step.
\newline
Because B-splines have local support, any trajectory segment can be represented by a fixed number of nearby control points and knots. Unlike discrete action chunks, however, a fixed number of control points does not imply a fixed time horizon because knot intervals may vary. The policy therefore predicts both the control points and knot values, outputting a fixed-size vector of local B-spline parameters, $[U; C]$, that can be used with standard imitation learning architectures with minimal modification.
\newline
In practice, evaluating a degree-$p$ B-spline segment requires neighboring control points and knots for boundary support. We append future control points when available and repeat the final control point near the end of a trajectory, producing fixed-size training targets while preserving valid B-spline evaluation.

\paragraph{High-frequency Execution and Temporal Rescaling.}
During inference, we decouple the policy inference rate from the low-level control rate. The visuomotor policy runs at a relatively low frequency and predicts a future B-spline segment. The low-level controller then samples this continuous segment at a much higher frequency to generate dense robot commands.

This design is important for fast manipulation. When executing a policy faster than the demonstration, discrete action chunks may become too sparse or discontinuous for reliable tracking. In contrast, a B-spline segment can be sampled densely after temporal rescaling, allowing the controller to receive smooth high-frequency commands even when the policy itself runs at a lower rate.

Because the predicted trajectory is continuous, we can directly adjust its execution speed by changing the mapping between real time and the spline parameter. Given a predicted trajectory $\mathbf{a}(u)$ and a target speedup factor $n$, the executed action is $\mathbf{a}_{\mathrm{exec}}(t) = \mathbf{a}(nt)$. This preserves the geometric shape of the predicted action trajectory while traversing it faster in real time.


\noindent\textbf{Pipelined Execution with Segment Alignment.}
For long-horizon tasks, the policy repeatedly predicts new B-spline segments while the robot is executing the previous one. A naive pipelined implementation assumes that the beginning of the newly predicted segment aligns with the unexecuted tail of the previous segment. In practice, this assumption is often violated due to policy prediction noise, inference latency, and controller tracking error. As a result, directly switching to the new segment can introduce discontinuities at segment boundaries.

To address this issue, we introduce an inference-time segment alignment mechanism. Let $a_{\text{last}}$ denote the most recently executed action, and let $S{\mathrm{new}}(t)$ denote the newly predicted B-spline segment evaluated at relative time $t$. Instead of starting the new segment at a fixed time determined only by inference latency, we search for the point on the new segment that best matches the current executed action:
\begin{equation}
\label{eq:align-time}
t^\star = \arg\min_{0 \leq t \leq \lambda T_{\mathrm{inf}}}
\mathrm{MSE}\big(S_{\mathrm{new}}(t), a_{\text{last}}\big),
\end{equation}

where $T^{\text{inf}}$ is the measured inference latency and $\lambda > 1$ slightly enlarges the search window for robustness. The controller then begins executing the new segment from $t^\star$.

This alignment step reduces boundary mismatch between consecutive B-spline segments and prevents small discontinuities from accumulating over time. As shown in our experiments, this becomes especially important under high-speed execution, where even small jumps in commanded actions can lead to tracking errors or task failure.

\vspace{-7pt}
\section{Experiments}
\vspace{-3pt}

We integrate \our with two imitation-learning backbones:
(i) Diffusion Policy~\cite{dp}, denoted as Diff.+BSP, and
(ii) Action Chunking with Transformers (ACT)~\cite{act}, denoted as Reg.+BSP.
We evaluate \our along two main metrics: 
\textbf{(1)} \textit{\textbf{success rate}}, measured across a diverse set of tasks, and 
\textbf{(2)} \textit{\textbf{average completion time}}, measured by the mean task completion time across successful rollouts.
To this end, we evaluate our method against baseline policies on three real-world tasks that span diverse manipulation challenges: \pickcube, which requires precise object manipulation; \cleantable, which involves long-horizon tabletop rearrangement; and \stackunstack, which emphasizes bimanual, contact-rich interaction. We also benchmark \our in simulation environments on \textit{Push-T}~\cite{florence2022implicit,dp}, \textit{RoboMimic}~\cite{robomimic2021} and \textit{RoboCasa}~\cite{nasiriany2024robocasa}.


\subsection{Real-world Experiments}
\label{sec:real-eval}
\vspace{-5pt}

\begin{figure*}[t]
    \centering
    \includegraphics[width=1\linewidth]{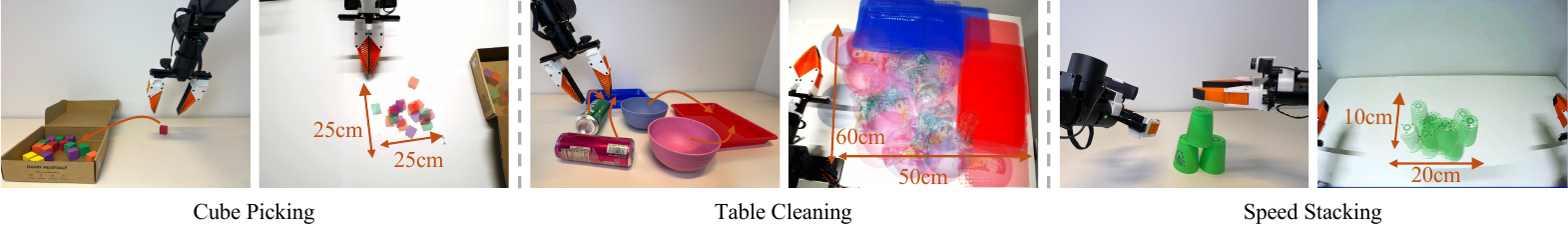}
    \vspace{-10pt}
    \caption{\textbf{Overview of the real-world tasks.} Task setups and corresponding object initialization distributions and workspace sizes for \pickcube, \cleantable, and \stackunstack.  These tasks vary in precision requirements, horizon length, and object interactions, highlighting the challenges of fast manipulation due to tight precision requirements, long task horizons, and complex object interactions.}

    \vspace{-10pt}

\label{fig:real-world-setup}
\end{figure*}

\noindent\textbf{Task setup.} In real-world experiments, we evaluate \our in the real world on three challenging tasks, as illustrated in Fig.~\ref{fig:real-world-setup}.
\noindent\textbf{\textit{1) \pickcube}.}
The robot is required to pick a cube and place it into a box on the table. We collect 200 demonstrations using cubes of four different colors. During both training and evaluation, only a single cube is present in the scene.
The cube has a side length of $3\,\mathrm{cm}$, requiring precise grasping and placement. \noindent\textbf{\textit{2) \cleantable}.}
This long-horizon manipulation task presents a significant challenge: the robot must clear a table by sequentially completing two subtasks: (i) picking two beverage cans and placing them into a designated plastic tray, and (ii) picking two empty bowls and placing them into a second tray, stacking the second bowl on top of the first. We collect 300 demonstrations for this task. 
\noindent\textbf{\textit{3) \stackunstack}.}
The robot must first construct a cup pyramid (2--1) from identical cups and then collapse it into a single nested stack using two arms.
Cups are randomly initialized within a clutter-free workspace. We collect 200 demonstrations and evaluate performance across diverse initial configurations.

All experiments use 6-DoF ARX5 robot arms. For \pickcube and \cleantable, we use a single third-person camera. For \stackunstack, we employ two wrist-mounted cameras in addition to a fixed third-person camera placed between the two arms.

\noindent\textbf{Baselines and Evaluation protocol.}
We compare \our against three classes of baselines:
\textbf{(1)} Diffusion policy~\cite{dp} and Regression policy~\cite{act}, which executes them at the original demonstration or control frequency. \textbf{(2)} Diffusion policy-$n$X and Regression policy-$n$X, which naively increase the execution frequency by a factor of $n$. This represents the most direct baseline for accelerating task execution. \textbf{(3)} DemoSpeedup~\cite{demospeedup}, which accelerates execution by temporally downsampling demonstration trajectories. 
For each real-world task, we evaluate all methods on 20 rollouts with distinct object layouts. To ensure fair comparisons, we first record a fixed set of test configurations using the camera system; for subsequent evaluations, the environment is reset by matching object placements to these recorded configurations. The distribution of all test configurations is shown in Fig.~\ref{fig:real-world-setup}. For grasping actions in \pickcube and \cleantable, we allow up to three grasp attempts as recovery. We report the average completion time, measured from the start at the home pose to the final gripper opening that marks task completion, averaged over all successful rollouts.




\begin{table*}[t]
  \centering
\vspace{-10pt}
    
  \fontsize{5.8pt}{6.8pt}\selectfont
  \renewcommand{\arraystretch}{0.92}
  \setlength{\tabcolsep}{1.2pt}

  \resizebox{\textwidth}{!}{
  \begin{tabular}{lc|cc|cc|cc||cc|cc|cc||ccc}
    \toprule
    Task & Metric
    & \multicolumn{6}{c||}{\textbf{Diffusion Policy}}
    & \multicolumn{6}{c||}{\textbf{Regression Policy}}
    & \multicolumn{3}{c}{\textbf{DemoSpeedUp Comparison}} \\
    \cmidrule(lr){3-8}
    \cmidrule(lr){9-14}
    \cmidrule(lr){15-17}

    & 
    & \multicolumn{2}{c|}{1X}
    & \multicolumn{2}{c|}{2X}
    & \multicolumn{2}{c||}{4X}
    & \multicolumn{2}{c|}{1X}
    & \multicolumn{2}{c|}{2X}
    & \multicolumn{2}{c||}{4X}
    & DemoSpeedUp
    & \makecell{Diff. +\\BSP 4X}
    & \makecell{Reg. +\\BSP 4X} \\
    \cmidrule(lr){3-4}
    \cmidrule(lr){5-6}
    \cmidrule(lr){7-8}
    \cmidrule(lr){9-10}
    \cmidrule(lr){11-12}
    \cmidrule(lr){13-14}

    & 
    & Diff.
    & \makecell{Diff. +\\BSP}
    & Diff.
    & \makecell{Diff. +\\BSP}
    & Diff.
    & \makecell{Diff. +\\BSP}
    & Reg.
    & \makecell{Reg. +\\BSP}
    & Reg.
    & \makecell{Reg. +\\BSP}
    & Reg.
    & \makecell{Reg. +\\BSP}
    & 
    & 
    & \\
    \midrule

    \makecell{Cube \\ Picking} & \cellMetric
    & \cellSRtime{19/20}{6.48s}
    & \cellSRtime{\greenNum{20/20}}{\purpleNum{6.31s}}
    & \cellSRtime{20/20}{4.58s}
    & \cellSRtime{\greenNum{20/20}}{\purpleNum{3.43s}}
    & \cellSRtime{20/20}{3.52s}
    & \cellSRtime{\greenNum{20/20}}{\purpleNum{2.45s}}
    & \cellSRtime{18/20}{9.84s}
    & \cellSRtime{\greenNum{19/20}}{\purpleNum{7.92s}}
    & \cellSRtime{15/20}{5.55s}
    & \cellSRtime{\greenNum{19/20}}{\purpleNum{3.36s}}
    & \cellSRtime{19/20}{3.74s}
    & \cellSRtime{\greenNum{19/20}}{\purpleNum{2.08s}}
    & \cellSRtime{18/20}{6.66s}
    & \cellSRtime{\greenNum{20/20}}{2.45s}
    & \cellSRtime{19/20}{\purpleNum{2.08s}} \\

    \midrule

    \makecell{Table \\ Cleaning} & \cellMetric
    & \cellSRtime{11/20}{41.40s}
    & \cellSRtime{\greenNum{15/20}}{\purpleNum{36.97s}}
    & \cellSRtime{\greenNum{15/20}}{28.87s}
    & \cellSRtime{14/20}{\purpleNum{20.81s}}
    & \cellSRtime{\greenNum{12/20}}{23.31s}
    & \cellSRtime{11/20}{\purpleNum{18.05s}}
    & \cellSRtime{13/20}{49.70s}
    & \cellSRtime{\greenNum{18/20}}{\purpleNum{38.47s}}
    & \cellSRtime{13/20}{34.48s}
    & \cellSRtime{\greenNum{18/20}}{\purpleNum{19.11s}}
    & \cellSRtime{13/20}{23.57s}
    & \cellSRtime{\greenNum{14/20}}{\purpleNum{11.80s}}
    & \cellSRtime{3/20}{49.47s}
    & \cellSRtime{11/20}{18.05s}
    & \cellSRtime{\greenNum{14/20}}{\purpleNum{11.80s}} \\

    \midrule

    \makecell{Speed \\ Stacking} & \cellMetric
    & \cellSRtime{14/20}{20.47s}
    & \cellSRtime{\greenNum{14/20}}{\purpleNum{18.13s}}
    & \cellSRtime{15/20}{13.66s}
    & \cellSRtime{\greenNum{15/20}}{\purpleNum{11.45s}}
    & \cellSRtime{\greenNum{9/20}}{10.49s}
    & \cellSRtime{7/20}{\purpleNum{9.62s}}
    & \cellSRtime{8/20}{19.75s}
    & \cellSRtime{\greenNum{16/20}}{\purpleNum{17.61s}}
    & \cellSRtime{4/20}{13.49s}
    & \cellSRtime{\greenNum{13/20}}{\purpleNum{10.98s}}
    & \cellSRtime{\greenNum{8/20}}{\purpleNum{10.42s}}
    & \cellSRtime{0/20}{NA}
    & \cellSRtime{3/20}{18.57s}
    & \cellSRtime{\greenNum{7/20}}{\purpleNum{9.62s}}
    & \cellSRtime{0/20}{NA} \\
    


    \bottomrule
  \end{tabular}
  }

  \caption{\textbf{Real-world quantitative results.}
  We report each method's success rate and average completion time on the real-world tasks.
  Completion time is averaged over successful rollouts. We highlight the results with higher success rates and shorter completion times in each comparison.}

  \vspace{-15pt}
  \label{tab:act-dp-results}
\end{table*}

\subsection{Real-world Results}
\vspace{-5pt}

Table~\ref{tab:act-dp-results} reports the overall success rates and average completion times. We organize our analysis around the following key findings.
\vspace{-5pt}

\textbf{Finding 1: BSP consistently shortens task completion time.} As highlighted in Tab.~\ref{tab:act-dp-results}, across all configurations, integrating BSP consistently reduces \textit{Avg. time}, regardless of whether it is paired with a Diffusion or Regression policy.
The efficiency of BSP acceleration is particularly pronounced in the long-horizon \cleantable task. Specifically, at a 4X speedup, integrating BSP with the Regression policy compresses the average completion time from $23.57\text{s}$ to $11.80\text{s}$ ($50\%$ reduction in duration), while simultaneously preserving task success rate ($13/20$ vs. $14/20$).

\textbf{Finding 2: BSP preserves task success, but aggressive speedup can exceed controller limits.}
\label{finding2}
BSP achieves faster execution while generally maintaining task success. Across diffusion and regression comparisons, BSP matches or improves success rate in 14 out of 18 settings.
This trend is particularly clear for the regression backbone: BSP improves the long-horizon Table Cleaning success from 13/20 to 18/20 at both $1\times$ and $2\times$ speedup, and improves Speed Stacking from 8/20 to 16/20 at $1\times$ speedup and from 4/20 to 13/20 at $2\times$ speedup.
\newline
At the same time, the results reveal a natural trade-off between execution speed and task robustness. 
%
For example, on \stackunstack, Regression+BSP outperforms the baselines at $1\times$ and $2\times$ speedup, but fails at $4\times$, achieving zero success because the aggressive BSP temporal scaling pushes the robot beyond the tracking limits of its low-level controller. 

\begin{figure*}[t]
    \centering
    \includegraphics[width=1\linewidth]{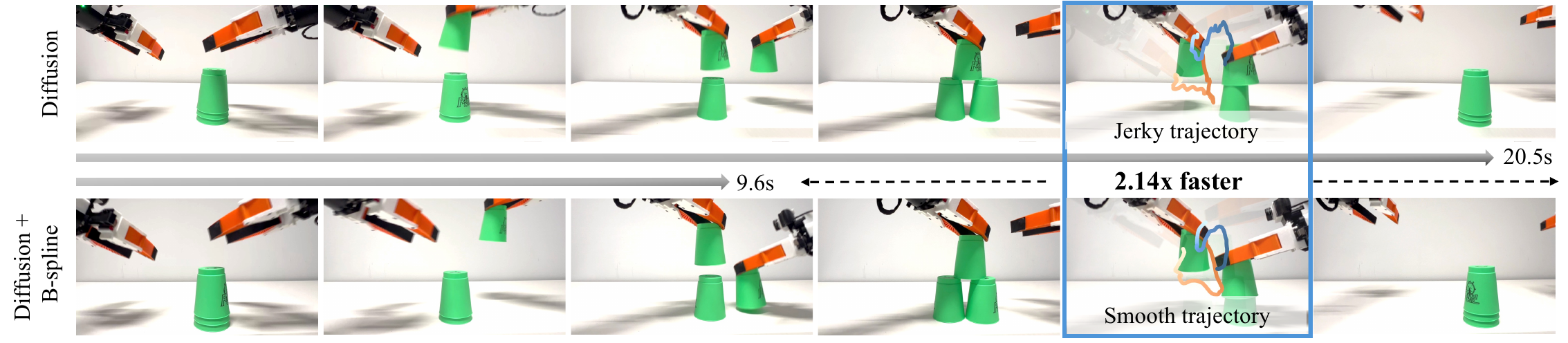}
\caption{\textbf{Real-world qualitative results.} We visualize key frames from the rollouts of \our and a diffusion policy on the speed stacking task, with the robot gripper tip trajectories overlaid on the frame, illustrating that \our achieves faster and smoother task execution.}
    \vspace{-15pt}

    \label{fig:qualitative-results}
\end{figure*}

\textbf{Finding 3: BSP produces smoother trajectories.}
During the experiments, we observed that \our produces substantially smoother executions than the baselines. We provide qualitative comparisons between \our and Diffusion Policy on the \stackunstack task in Fig.~\ref{fig:qualitative-results}, where we overlay the executed trajectories on video frames. \our yields smoother and faster motions, while the action chunks produced by Diffusion Policy often result in jerky executions due to discontinuities between consecutive chunks. Please visit \href{https://b-spline-policy.github.io/}{our website} for video results.

%


\begin{wrapfigure}{r}{0.45\textwidth}
  \centering
  \vspace{-10pt}
  \includegraphics[width=\linewidth]{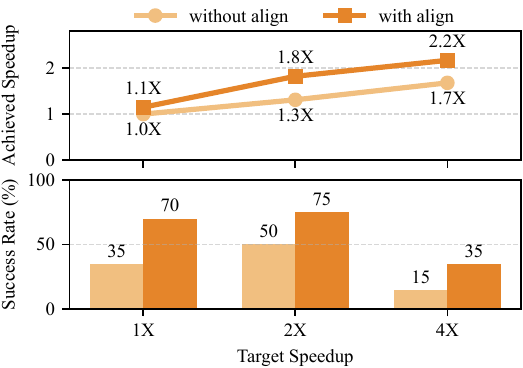}
    
  \caption{\textbf{Ablation study on inference-time segment alignment.}
  Aligning adjacent segments improves trajectory continuity, which becomes critical at high speeds, leading to higher success rates and more consistent realized speedups.}
  \label{fig:infer-align-success-speed}
  \vspace{-10pt}
\end{wrapfigure}

\textbf{Finding 4: Inference-time segment alignment is critical for BSP.}
Consistent with the smoother trajectories observed in Finding 3, we further find that trajectory continuity across consecutive spline segments is critical for stable high-speed execution. To better understand this, we conduct an ablation study on inference-time segment alignment on \stackunstack task. 
As shown in Fig.~\ref{fig:infer-align-success-speed}, segment alignment becomes increasingly important at higher execution speeds: when continuity is not enforced, success becomes highly sensitive to execution speed. From the policy perspective, such discontinuities effectively induce out-of-distribution (OOD) inputs, forcing the policy to spend additional steps replanning to return to the in-distribution regime, or causing outright task failure.

\vspace{-10pt}
\subsection{Simulation Benchmarks}
\vspace{-5pt}

For reproducibility, we also evaluate \our on simulated manipulation tasks to assess whether continuous B-spline action representations achieve performance comparable to standard action-chunking imitation learning methods on \textit{Push-T}~\cite{florence2022implicit,dp}, \textit{RoboMimic}~\cite{robomimic2021} and \textit{RoboCasa}~\cite{nasiriany2024robocasa}.

\begin{table*}[t]
  \centering
  \vspace{-6pt}

  \captionsetup[subtable]{
    justification=centering,
    singlelinecheck=false,
    font=footnotesize,
    skip=2pt
  }

  \begin{subtable}[t]{0.58\textwidth}
    \begin{minipage}[t][0.26\textheight][t]{\linewidth}
    \centering
    \fontsize{6.3pt}{7.3pt}\selectfont
    \renewcommand{\arraystretch}{0.92}
    \setlength{\tabcolsep}{2.0pt}

    \resizebox{\linewidth}{!}{
    \begin{tabular}{l|cccccc}
      \toprule
      Configuration 
      & PushT 
      & \makecell{Lift} 
      & \makecell{Turn off\\sink faucet} 
      & \makecell{Coffee\\press button} 
      & \makecell{Turn off\\microwave} 
      & \makecell{Close\\door} \\
      \midrule

      Diff. 1X Base  & 72\%            & 100\%             & 79\%            & 93\%            & 77\%            & 27\% \\
      Diff. 1X +BSP  & \greenNum{75\%} & \greenNum{100\%}  & \greenNum{85\%} & \greenNum{94\%} & \greenNum{89\%} & \greenNum{46\%} \\
      \midrule
      Reg. 1X Base   & 63\%            & \greenNum{99\%}   & 84\%            & 89\%            & 93\%            & 40\% \\
      Reg. 1X +BSP   & \greenNum{66\%} & 91\%              & \greenNum{94\%} & \greenNum{93\%} & \greenNum{97\%} & \greenNum{60\%} \\

      \bottomrule
    \end{tabular}
    }

    \caption{\textbf{Sim results.}}
    \label{tab:sim-results}
    \end{minipage}
  \end{subtable}
  \hfill
  \begin{subtable}[t]{0.39\textwidth}
    \begin{minipage}[t][0.26\textheight][t]{\linewidth}
    \centering
    \fontsize{5.8pt}{6.8pt}\selectfont
    \renewcommand{\arraystretch}{0.88}
    \setlength{\tabcolsep}{1.4pt}

    \resizebox{\linewidth}{!}{
    \begin{tabular}{c|cc|cc|cc}
      \toprule
      Metric
      & \multicolumn{6}{c}{\textbf{Diffusion Policy}} \\
      \cmidrule(lr){2-7}

      &
      \multicolumn{2}{c|}{1X}
      & \multicolumn{2}{c|}{2X}
      & \multicolumn{2}{c}{4X} \\
      \cmidrule(lr){2-3}
      \cmidrule(lr){4-5}
      \cmidrule(lr){6-7}

      &
      Diff.
      & \makecell{Diff. +\\BSP}
      & Diff.
      & \makecell{Diff. +\\BSP}
      & Diff.
      & \makecell{Diff. +\\BSP} \\
      \midrule

      \pushtcellMetric
      & \cellSRtime{0.47}{12.24s}
      & \cellSRtime{\greenNum{0.65}}{\purpleNum{7.20s}}
      & \cellSRtime{0.57}{9.22s}
      & \cellSRtime{\greenNum{0.71}}{\purpleNum{4.89s}}
      & \cellSRtime{0.59}{7.19s}
      & \cellSRtime{\greenNum{0.73}}{\purpleNum{2.87s}} \\

      \bottomrule
    \end{tabular}
    }

    \caption{\textbf{PushT-speedup results.}}
    \label{tab:pusht}
    \end{minipage}
  \end{subtable}

  \vspace{-90pt}
  \caption{\textbf{Simulation results.} In Table (a), we report average score on pushT and average success rate across other simulation benchmarks.
  In Table (b), we report the average score and average time for the PushT speedup.}
  \label{tab:combined-sim-real-results}
  \vspace{-10pt}
\end{table*}

\noindent\textbf{Quantitative performance.}
As shown in Table~\ref{tab:sim-results}, \our matches or outperforms the corresponding base policies across all evaluated tasks, except the RoboMimic Lift task.
%
On the more challenging RoboCasa tasks, \our consistently outperforms the baseline methods, demonstrating improved robustness in long-horizon manipulation scenarios.
Overall, these results indicate that representing actions as B-spline trajectories does not compromise imitation learning performance and can yield consistent gains over discrete-time action-chunking representations. 

\noindent\textbf{Acceleration in simulation.} Accelerating policies in simulation requires modifying the environment and low-level controllers to support the high-frequency execution required by BSP, which then requires re-collecting datasets, making it difficult to directly measure policy acceleration in each simulation benchmark. Therefore, we conduct this study on Push-T, a representative task that is easy to modify.
We collect 100 demonstrations in the PushT-speedup environment at 200 Hz. BSP is trained directly on the 200 Hz trajectories, while the diffusion policy baseline is trained on 10 Hz trajectories downsampled from the same data.
For each checkpoint, we evaluate 50 rollouts and report the average score. We define a rollout as successful if its score exceeds 0.9, which indicates near-complete coverage. For each successful rollout, we record the first time step at which the score exceeds 0.9, and report the average of this time across successful rollouts. For each setting, we evaluate the best three checkpoints and report the mean of these metrics.\newline 
As shown in Table~\ref{tab:pusht}, BSP consistently improves both the average score and completion time across all target speedups. The gains become pronounced at higher speedups: at $4\times$, Diff.+BSP achieves a higher average score ($0.73$ vs. $0.59$) while reducing completion time by more than half ($7.19$s to $2.87$s). This is because faster execution enables the Push-T agent to perform more corrective pushes within the maximum rollout time, leading to higher final coverage scores. 
%




\vspace{-10pt}
\section{Limitations}
\vspace{-10pt}
\paragraph{Hardware Constraints and Policy Acceleration Limits.} As detailed in Section~\ref{finding2}, aggressively accelerating the policy introduces a drop in robustness; specifically, the Speed Stacking task drops to a 0\% success rate under a 4x speedup. 
This failure is primarily driven by the physical constraints of the low-cost robotic arms used in our experiments, whose low-level controllers lack the stiffness and accuracy required to track rapid action commands. We believe that integrating improved low-level controllers would significantly improve tracking robustness at extreme speeds, which we plan to explore in future work.

\vspace{-10pt}
\section{Conclusion}
\vspace{-10pt}
We propose \our, a novel action representation that parameterizes the actions as continuous curves using B-splines. \our improves trajectory smoothness and enables faster task completion with robust success rate.
To ensure stability during high-speed execution, we propose an inference-time segment alignment mechanism that effectively mitigates boundary discontinuities between successive spline segments.
Extensive empirical evaluations across diverse simulated and real-world manipulation tasks demonstrate that BSP significantly reduces task completion times, while consistently preserving or enhancing success rates over standard baselines.

\clearpage


\bibliography{example}  

\clearpage
\appendix
\section{Appendix}
\subsection{Implement Details}

\noindent\textbf{B-spline action representation. }
We represent action trajectories using cubic B-splines with degree $k=3$.
In simulation, we follow standard benchmark settings and use a delta end-effector action space.
In real-world experiments, we use absolute joint-position actions.

\noindent\textbf{Adaptive b-spline fitting from demonstrations. }
We fit a B-spline to each demonstration trajectory using the adaptive knot insertion procedure in Alg.~\ref{alg:spline-compress}.
For most tasks, we use a small fitting tolerance $\varepsilon=0.002$ to preserve high-precision behaviors.
For Push-T, we use $\varepsilon=1$ because the action magnitude is substantially larger than in the other benchmarks, with values ranging roughly from $0$ to $512$.

\noindent\textbf{Model training of \our.} 
We train \our model to predict a fixed-size parameter segment consisting of 16 knots and their corresponding control points, including six boundary-support knots.
Because observations are not aligned to knot or control-point indices, at each observation step, the model predicts the nearest future segment containing the next 16 knots and control points.
In simulation, all models use the RoboMimic visual encoder.
In real-world experiments, diffusion uses the same visual encoder, whereas the regression uses a DINOv2 visual encoder.

\noindent\textbf{Model training of action-chunk policies.} 
For a fair comparison, all action-chunk baseline policies (Diffusion/Regression) are also trained to predict 16 future actions.
Each baseline is trained for the same number of epochs as its B-spline counterpart.
The visual encoder is the same as \our.

\noindent\textbf{DemoSpeedup.} We implement the DemoSpeedup based on the diffusion policy. 
The method uses two key hyperparameters, which are the downsampling rates for the precision and non-precision phases.
We found that downsampling the precision phase by a factor $\ge 2$ prevents the policy from completing the task reliably.
We therefore set the precision-phase downsampling rate to 1.
For the non-precision phase, we use a downsampling rate of 4 for \pickcube\ and \cleantable.
We use 2 for \stackunstack\ because it requires higher accuracy.

\noindent\textbf{Controller.}
In simulation, actions are delta end-effector commands.
We sample B-spline actions at 100\,Hz, but we do not apply segment alignment because it does not have a clear physical interpretation for delta actions.
In real-world experiments, actions are absolute joint positions.
We sample actions from each predicted B-spline segment at 100\,Hz.
For phase alignment, we set the search window to $[0,\lambda T^{\mathrm{inf}}]$ with $\lambda T^{\mathrm{inf}} = u_{\max}(S)/2$, where $u_{\max}(S)$ denotes the segment horizon in the spline parameter domain.

\noindent\textbf{Knot validity projection.}
A valid B-spline requires a nondecreasing knot vector, which standard imitation-learning regressors do not enforce. In practice predicted knots are almost always monotonic; for any knot $u_i$ that violates the constraint we apply $u_i \leftarrow \max(u_i,\, u_{i-1} + \delta)$ with a small constant $\delta > 0$, restoring validity with minimal perturbation.

\noindent\textbf{Inference techniques.} During inference, pipelined execution with segmentment alignment is applied. Here we detailed the complete procedure in Algorithm~\ref{alg:pipeline-inference}.

\begin{table}[h]
\centering
\setlength{\tabcolsep}{10.0pt}
\begin{tabular}{c|cc}
\toprule
Push-T & Score & Compression Ratio \\
\midrule
$\varepsilon{=}0.5$ & 0.52 & 1.12 \\
$\varepsilon{=}1$   & 0.65 & 1.34 \\
$\varepsilon{=}2$   & 0.62 & 1.73 \\
$\varepsilon{=}4$   & 0.63 & 2.36 \\
$\varepsilon{=}8$   & 0.65 & 3.34 \\
\bottomrule
\end{tabular}
\caption{\textbf{Ablation on fitting error tolerance.}
We report the success rate and the compression ratio under different maximum fitting errors $\varepsilon$.}
\label{tab:pushT_fit_error}
\end{table}

\begin{algorithm}[h]
\caption{Inference pipeline}
\label{alg:pipeline-inference}
\textbf{Input:} policy $\pi$, speedup factor $m$, inference budget $\delta^{\text{budget}}$ \\
\textbf{State:} current B-spline segment $S$, segment start time $t_0$, in-flight inference flag \quad in\_flight
\begin{enumerate}[leftmargin=1.2em]
\item \textbf{Initialize:} $S \leftarrow \pi(o_{\text{recent}})$,\quad $t_0 \leftarrow \text{now}()$,\quad in\_flight $\leftarrow$ false.
\item \textbf{Loop (control at high rate):}
\begin{enumerate}[leftmargin=1.2em]
\item $t_{\text{real}} \leftarrow \text{now}() - t_0$, \quad $u \leftarrow m \cdot t_{\text{real}}$ \hfill(\textit{Speedup})
\item Execute $a \leftarrow S(u)$ \hfill(\textit{Sample})
\item \textbf{If} $(u_{\max}(S) - u) < \delta^{\text{budget}}$ and in\_flight = false:
\begin{enumerate}[leftmargin=1.2em]
\item Launch asynchronous inference to obtain a new segment $S_{\text{new}} \leftarrow \pi(o_{\text{recent}})$
\item Set in\_flight $\leftarrow$ true
\end{enumerate}
\item \textbf{On inference completion:} let $T^{\text{inf}}$ be the measured inference latency.
\begin{enumerate}[leftmargin=1.2em]
\item Compute alignment time $t^\star$ by Eq.~\eqref{eq:align-time} (using $S_{\text{new}}$ and $a_{\text{last}}$)
\item Swap $S \leftarrow S_{\text{new}}$
\item Update $t_0 \leftarrow \text{now}() - \frac{t^\star}{m}$ \hfill(\textit{Segment align})
\item Set in\_flight $\leftarrow$ false
\end{enumerate}
\end{enumerate}
\end{enumerate}
\end{algorithm}

\subsection{Additional experiments results}

\noindent\textbf{Ablation on fitting error tolerance.}
Representing actions with B-splines introduces a fitting tolerance $\varepsilon$.
This typically yields fewer knot and control point pairs than the original demonstrations.
Tab.~\ref{tab:pushT_fit_error} reports success rate and compression ratio under different tolerances.
We define the compression ratio as the number of original actions divided by the number of knot--control-point pairs.
As $\varepsilon$ increases, the compression ratio increases substantially while success rate remains relatively stable.
This suggests that performance is not highly sensitive to $\varepsilon$ within a reasonable range, and a moderately larger tolerance can reduce computational cost.

\end{document}